\documentclass{article}

\PassOptionsToPackage{numbers, compress}{natbib}
\usepackage[final]{neurips_2022}
\usepackage{multirow}

\usepackage[utf8]{inputenc} 
\usepackage[T1]{fontenc}    
\usepackage{hyperref}       
\usepackage{url}            
\usepackage{booktabs}       
\usepackage{amsfonts}       
\usepackage{nicefrac}       
\usepackage{microtype}      
\usepackage{xcolor}         
\usepackage{graphicx}
\usepackage{float}
\usepackage{caption}
\usepackage{subcaption}
\usepackage{amsmath}
\usepackage{amssymb}
\usepackage{array, booktabs, makecell}
\usepackage{csquotes}
\usepackage{setspace}

\title{Towards Reliable Zero Shot Classification in Self-Supervised Models with Conformal Prediction}

\author{%
  Bhawesh Kumar\thanks{Denotes equal contribution} \\
  Department of Biostatistics\\
  Harvard T.H. Chan School of Public Health\\
  \texttt{bhawesh\_kumar@hsph.harvard.edu} \\
  \And
  Anil Palepu\textsuperscript{*} \\
  Harvard-MIT Health Sciences \& Technology \\
  Massachusetts Institute of Technology \\
  \texttt{apalepu@mit.edu} \\
  \AND
  Rudraksh Tuwani \\
  Department of Epidemiology\\
  Harvard T.H. Chan School of Public Health\\
  \texttt{rtuwani@hsph.harvard.edu} \\
  \And
  Andrew Beam \\
  Department of Epidemiology\\
  Harvard T.H. Chan School of Public Health\\
  \texttt{andrew\_beam@hms.harvard.edu} \\
}

\everypar{\looseness=-1}
\begin{document}

\maketitle

\begin{abstract}
  Self-supervised models trained with a contrastive loss such as CLIP have shown to be very powerful in zero-shot classification settings. However, to be used as a zero-shot classifier these models require the user to provide new captions over a fixed set of labels at test time. In many settings, it is hard or impossible to know if a new query caption is compatible with the source captions used to train the model. We address these limitations by framing the zero-shot classification task as an outlier detection problem and develop a conformal prediction procedure to assess when a given test caption may be reliably used. On a real-world medical example, we show that our proposed conformal procedure improves the reliability of CLIP-style models in the zero-shot classification setting, and we provide an empirical analysis of the factors that may affect its performance.   
\end{abstract}

\section*{Introduction}
\looseness=-1
 Self-supervised models such as CLIP \citep{radford2021learning} have demonstrated promising capabilities on zero-shot classification tasks, and have been shown to be competitive with fully supervised models in some settings \citep{radford2021learning, zhang2020contrastive, palepu2022self}. To perform zero-shot classification for a test image, CLIP-style models require the user to provide a set of query captions. Using this query set, the model provides the relative probabilities that each caption in the query set matches the test image of interest. This presents two immediate challenges: 1) The user must know \emph{a priori} the full set of labels to consider, and 2) the user must be able to write query captions in a similar manner to the captions used to train the base model. 
 
For example, consider a CLIP-style model trained on the task of matching chest X-rays (CXRs) to the corresponding CXR report \citep{zhang2020contrastive, tiu2022expert, palepu2022self}, which was the original inspiration for CLIP \citep{tiu2022expert}. Once trained, such a model can be used to predict disease status in new CXRs by querying the model with captions that describe the conditions of interest. However, the training captions were most likely written by radiologists using fine-grained, medical terminology that might be difficult for a non-specialist to replicate. Consider the following CXR report, taken from \citep{cohen2020covid}
\begin{displayquote}
FINDINGS: The lungs are clear without effusion, consolidation, or edema. Moderate cardiac enlargement and tortuosity of the descending thoracic aorta are again noted. Left shoulder arthroplasty changes in degenerative changes at the right ac joint are seen. Surgical clips project over the upper abdomen.\\
 
IMPRESSION: Cardiomegaly without superimposed acute cardiopulmonary process.
\end{displayquote}
This patient has several outcomes that would be of interest for zero-shot classification. The first sentence in the FINDINGS section indicates that they do \emph{not} have the conditions of lung opacity, pleural effusion, consolidation, or edema. The second sentence indicates that the patient \emph{does} have cardiomegaly (an enlarged heart). Thus, to perform zero-shot classification on new CXRs, the user must write sufficiently rich caption for these conditions so that the base model is able to align the caption with the query image. There is currently no mechanism for users of CLIP-style models to know if they have composed good query captions. 

To address these issues, we propose a new framework for contrastive self-supervised models that seeks to answer two questions: 1) Is the similarity score between a query caption and query image sufficiently high relative to the distribution of scores seen in the training set, and 2) Is the query caption sufficiently similar to the captions used to train the model? To answer these questions, we adopt a conformal prediction perspective \citep{angelopoulos2021gentle, shafer2008tutorial} that quantifies the extent to which a query caption is an \emph{outlier} according to these two criteria. Conformal prediction \citep{shafer2008tutorial} is a statistical framework that provides uncertainty quantification with statistical guarantees for a variety of scenarios with minimal assumptions \citep{kompa2021second, angelopoulos2021gentle}. Conformal methods rely on a non-conformity measure to assess how well a test input ``conforms'' with previously seen data points. 

For the purpose of quantifying uncertainty of query captions, we propose two novel non-conformity scores. First, we define a non-conformity score based on the cosine distance between a paired caption $x$ and an image $y$:
\begin{align}
    s_I(x, y) = 1 - \frac{x \cdot y}{||x||_2||y||_2}
    \label{eqn:sI}
\end{align}
Next, we define a non-conformity score based on the mean k-nearest neighbors distance (KNND) between a caption $x$ and the set of captions $X$ used to train the model:
\begin{align}
    s_T(x, X) = \texttt{KNND}(x,X)
    \label{eqn:sT}
\end{align}
where the base KNND metric is again cosine distance. $s_I$ measures how well a given query caption matches a query image relative to those seen during training while $s_T$ measures how similar the query caption is to the captions used to train the model. We estimate the empirical quantities of $s_I$ and $s_T$ using a held-out calibration set. We declare a query caption to be an outlier if its score $s_T$ is less than $\alpha$ quantile (e.g. 0.05) of the empirical conformal score distribution. Under the assumptions of exchangability, the following guarantee on the error rate will hold (see section 4.4 of \citep{angelopoulos2021gentle}):
\begin{align}
    \mathbb{P}(\mathcal{C}(x_{query}) = \texttt{outlier}) \leq \alpha
    \label{eqn:error}
\end{align}
where $\mathcal{C}(x_{query}) = \mathtt{outlier}$ indicates $x_{query}$ is a member of the set of outlier points. Note that equation \ref{eqn:error} is a control for the \textbf{True Positive Rate (TPR)} in our zero-shot classification scheme using $s_I$, since we are admitting true positives with a pre-specified probability of $\alpha$. 

In summary, this work makes the following contributions:
\begin{itemize}
  \item Two new scores that assess the the compatibility of query captions with the base self-supervised model. One score measures the compatibility between the query caption and the test image, while the other score measures compatibility between the query caption the set of training captions.
  \item A conformal procedure to quantify uncertainty of these scores, and an empirical study of the properties of training captions that impact zero-shot classification performance.
\end{itemize}
\section*{Methods}

\subsection*{Data, Clip-style models, and Zero-shot Classification}
We used 243,324 frontal x-rays with corresponding radiology reports from the MIMIC-CXR-JPG \cite{johnson2019mimic} dataset, as well as labels indicating the presence or absence of various clinical findings. We focused on the Cardiomegaly, Edema, Consolidation, and Pleural Effusion labels in this work. Additionally, we parsed the radiology reports to extract the ``findings'' and ``impression'' sections \cite{liu2019clinically}.

We trained two CLIP-style \cite{radford2021learning} models: a ``Findings+Impressions model'' in which the training text was a concatenation of the findings and impression sections of the radiology reports, and the ``Impressions-only model'' in which only the impression section was provided during training. Further details about the modified CLIP architecture and training procedures are available in \ref{modelinfo}. We constructed zero-shot classifiers as in \cite{palepu2022self} to predict the four clinical labels. Further details about the zero-shot classification procedure and label queries are available in \ref{zeroshot} and \ref{query_caption}. 

\subsection*{Evaluation of Conformal Error Rates}
 We assessed the non-conformity score for images and text $s_I$ as described in equation (\ref{eqn:sI}). First, we computed the cosine distance between true image-text pairs in the calibration set, shown in figure \ref{fig:pos_neg_img_fig} (first row) of the appendix. Next, we computed the True Positive Rates (TPRs) and False Positive Rates (FPRs) by thresholding the cosine-distance of the query-image pairs at errors rates of $\alpha \in \{0.001, 0.01, 0.05, 0.1\}$. These error rates  aimed to control for the TPR of our label captions according to \ref{eqn:error}. We evaluated the TPR and FPR at these error rates for both models (Impressions-only and Findings+Impressions) and this was simulated 100 times using 5000 randomly selected conformal scores from the calibration set. 
 
 To assess the text non-conformity scores $s_T$ in equation \ref{eqn:sT}, we first calculated the mean cosine distance to the 500 nearest training text embeddings for each calibration set caption and obtained a distribution of KNND conformal scores (ref. figures \ref{fig:knn_conformal_i}, \ref{fig:knn_conformal_f} in the appendix). We obtained the conformal scores corresponding to the query captions in same way, and then we calculated error rate required to admit a query caption based on this non-conformity score. A lower error rate (or higher coverage rate) to admit a query caption was indicative that the query was less similar to the training and calibration captions.
\begin{table}[btp]
\small
 \caption{TPR and FPR for Mixed Labels at Different Conformal Coverage Rates}
  \centering
  \resizebox{\textwidth}{!}{%
    \begin{tabular}{cccccc}
    \toprule
    {\thead{Label}} & {\thead{Conformal Error Rate}} & {\thead{Findings + Impressions TPR}} &  {\thead{Impressions TPR}} & {\thead{Findings + Impressions FPR}} &  {\thead{Impressions FPR}}\\
    \midrule
    \multirow{4}{*}{Cardiomegaly} &0.001 & $0.607 \pm .033$ & $0.796 \pm .041$ & $0.225 \pm .024$& $0.480 \pm .062$ \\ 	
    &0.01 & $0.421 \pm .012$ & $0.541 \pm .016$ & $0.111 \pm .005$& $0.208 \pm .015$ \\ 
    &0.05 & $0.267 \pm .006$ & $0.365 \pm .009$ & $0.043 \pm .001$& $0.092 \pm .002$ \\ 
    &0.10 & $0.196 \pm .004$ & $0.284 \pm .006$ & $0.025 \pm .001$& $0.057 \pm .002$ \\ \midrule
    \multirow{4}{*}{Edema} &0.001 & $0.683 \pm .034$ & $0.970 \pm .017$ &$0.351 \pm .036$& $0.715 \pm .062$ \\ 	
    &0.01 & $0.484 \pm .017$ & $0.796 \pm .021$ &$0.193 \pm .011$& $0.397 \pm .023$ \\ 
    &0.05 & $0.302 \pm .007$ & $0.557 \pm .012$ &$0.088 \pm .003$& $0.190 \pm .007$ \\ 
    &0.10 & $0.221 \pm .005$ & $0.430 \pm .007$ &$0.054 \pm .002$& $0.114 \pm .003$ \\ \midrule
    \multirow{4}{*}{Consolidation} &0.001 & $0.484 \pm .049$ & $0.698 \pm .061$ & $0.266 \pm .042$& $0.234 \pm .070$ \\ 	
    &0.01 & $0.264 \pm .014$ & $0.431 \pm .019$ & $0.088 \pm .011$& $0.066 \pm .005$ \\ 
    &0.05 & $0.146 \pm .004$ & $0.234 \pm .008$ & $0.025 \pm .001$& $0.020 \pm .003$ \\ 
    &0.10 & $0.104 \pm .002$ & $0.169 \pm .003$ & $0.013 \pm .001$& $0.007 \pm .001$ \\ \midrule
    \multirow{4}{*}{Pleural Effusion} &0.001 & $0.928 \pm .017$ & $0.986 \pm .008$ &$0.678 \pm .064$& $0.967 \pm .018$ \\ 	
    &0.01 & $0.753 \pm .018$ & $0.844 \pm .020$ &$0.309 \pm .024$& $0.543 \pm .057$ \\ 
    &0.05 & $0.486 \pm .013$ & $0.500 \pm .017$ &$0.092 \pm .006$& $0.110 \pm .009$ \\ 
    &0.10 & $0.329 \pm .008$ & $0.299 \pm .015$ &$0.039 \pm .002$& $0.044 \pm .002$ \\ 
    \bottomrule
    \end{tabular}
    }
  \label{tab:table1}
  \small
 \caption{TPR and FPR for Pure Labels at Different Conformal Coverage Rate}
  \centering
  \resizebox{\textwidth}{!}{%
      \begin{tabular}{cccccc}
        \toprule

        {\thead{Label}} & {\thead{Conformal Error Rate}}    & {\thead{Findings + Impressions TPR}} &  {\thead{Impressions TPR}} & {\thead{Findings + Impressions FPR}} &  {\thead{Impressions FPR}}\\
        \midrule
        \multirow{4}{*}{Cardiomegaly} &0.001 & ${0.660 \pm .028}$ & ${0.874 \pm .027}$ & ${0.231 \pm .026}$& ${0.477 \pm .065}$ \\ 	
        &0.01 & ${0.482 \pm .014}$ & ${0.703 \pm .014}$ & ${0.112 \pm .006}$& ${0.210 \pm .017}$ \\ 
        &0.05 & ${0.308 \pm .007}$ & ${0.532 \pm .013}$ & ${0.043 \pm .001}$& ${0.092 \pm .002}$ \\ 
        &0.10 & ${0.226 \pm .003}$ & ${0.439 \pm .009}$ & ${0.025 \pm .001}$& ${0.057 \pm .002}$ \\ \midrule
        \multirow{4}{*}{Edema} &0.001 & ${0.800 \pm .027}$ & ${0.978 \pm .010}$ &${0.362 \pm .039}$& ${0.723 \pm .063}$ \\ 	
        &0.01 & ${0.624 \pm .015}$ & ${0.878 \pm .017}$ &${0.194 \pm .011}$& ${0.396 \pm .023}$ \\ 
        &0.05 & ${0.411 \pm .008}$ & ${0.691 \pm .011}$ &${0.088 \pm .003}$& ${0.190 \pm .006}$ \\ 
        &0.10 & ${0.307 \pm .008}$ & ${0.572 \pm .007}$ &${0.054 \pm .002}$& ${0.114 \pm .003}$ \\ \midrule
        \multirow{4}{*}{Consolidation} &0.001 & ${0.360 \pm .039}$ & ${0.585 \pm .055}$ & ${0.255 \pm .035}$& ${0.230 \pm .060}$ \\ 	
        &0.01 & ${0.171 \pm .005}$ & ${0.330 \pm .016}$ & ${0.090 \pm .009}$& ${0.066 \pm .005}$ \\ 
        &0.05 & ${0.094 \pm .004}$ & ${0.170 \pm .004}$ & ${0.025 \pm .001}$& ${0.020 \pm .003}$ \\ 
        &0.10 & ${0.073 \pm .004}$ & ${0.129 \pm .001}$ & ${0.013 \pm .001}$& ${0.006 \pm .001}$ \\ \midrule
        \multirow{4}{*}{Pleural Effusion} &0.001 & ${0.912 \pm .016}$ & ${0.985 \pm .008}$ &${0.666 \pm .061}$& ${0.966 \pm .019}$ \\ 	
        &0.01 & ${0.731 \pm .018}$ & ${0.867 \pm .017}$ &${0.309 \pm .023}$& ${0.539 \pm .051}$ \\ 
        &0.05 & ${0.459 \pm .012}$ & ${0.531 \pm .017}$ &${0.092 \pm .006}$& ${0.112 \pm .009}$ \\ 
        &0.10 & ${0.308 \pm .008}$ & ${0.354 \pm .017}$ &${0.038 \pm .002}$& ${0.044 \pm .003}$ \\ 
        \bottomrule
      \end{tabular}
      }
  \label{tab:table2}
\end{table}
\section*{Results}
\label{headings}
\subsection*{Control of the TPR using the image-caption score $s_I$}
We first assessed how well the conformal procedure for the image-caption non-conformity score $s_I$ controls the TPR. Table \ref{tab:table1} shows the TPR and FPR for models trained on different source captions. Neither model achieves the advertised TPR (i.e. the error rate controlled by the conformal procedure) of $\alpha$ in any of the tested scenarios. However, the Impressions-only model is closer to the correct error rate than the Findings+Impressions model. This is likely due to the fact that the impression sections more closely resembles the query captions in the sense that they concisely describe the patient's condition.

\subsection*{Effect of source caption complexity on the TPR}
We next evaluated the effect of caption complexity on control of the conformal error rate. Since our query captions only contain a description of single condition (see appendix), it is possible that looking at the distribution of all image-caption pairs is inappropriate given that many of the training captions contain descriptions of several conditions. We thus estimated the empirical distribution of non-conformity scores for single condition captions (``pure labels'') and compared this to the original distribution of multi-condition captions (``mixed labels''). In this scenario, we found that that using the pure label scores generally resulted in better TPRs when compared mixed label images across both the Impressions-only and Findings+Impressions models as seen in Table \ref{tab:table2}.

\subsubsection*{Caption query distance $s_T$ identifies reliable zero-shot tasks}
\begin{figure}[H]
    \centering
    \includegraphics[scale=0.45]{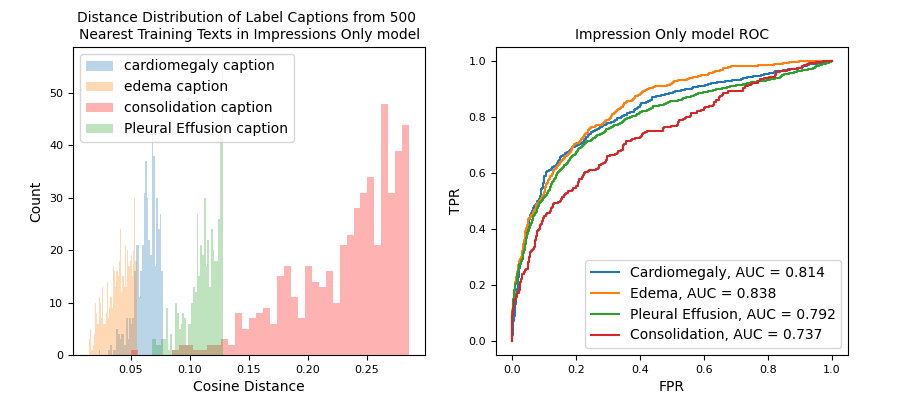}
    \caption{Left plot shows histograms of cosine distances from each label caption embedding to their 500 nearest text embeddings in the train set for Impressions-only model. On the right are the ROC curves for these label captions corresponding to the zero-shot classification on test set with pure labels.}
    \label{fig:ROC_KNN_I}
\end{figure}

For the final experiment we analyzed if the KNN distance between a query caption and the training captions, $s_T$,
could identify which zero-shot tasks were reliable. We found that a larger $s_T$ for a label caption was indicative of worse classification performance for that label as seen in figure \ref{fig:ROC_KNN_I} and \ref{fig:ROC_KNN_IF}. Edema, with the lowerst average $s_T$, had the best AUROC score, while Consolidation, with the highest average $s_T$, had the lowest AUROC score.
\section*{Discussion}

In this work we propose a new method for assessing the reliability of zero-shot contrastive self-supervised models. Building on the observation that constructing appropriate query captions is non-trivial in complex domains such as medicine, we provide a principled method for assessing the suitability of query captions relative to those used to train the base model. Further, we propose the first use of a conformal procedure for CLIP-style models that attempts to admit realistic captions with statistical guarantees. We show that the naive conformal procedure fails to control the error rate at the advertised level, and is sensitive to both the type of text used to train the model as well as the presence of additional findings in test samples, pointing to a violation of the exchangability assumption needed for conformal methods. However, the proposed method provides a new and valuable heuristic for assessing caption quality that is predictive of zero-shot performance on downstream tasks.

\newpage
\bibliography{references.bib}

\newpage
\appendix

\section{Appendix}

\subsection{Model architecture and training procedure} \label{modelinfo} 
We created train, validation, calibration, and test data sets from MIMIC-CXR with the following proportions: 0.72, 0.08, 0.1, 0.1. These sets were stratified such that a patient would never be represented in more than one set even if they had multiple studies or CXRs within the dataset. The train and validation sets were used to train the model, the calibration set was used to compute conformal scores, and the test set was used to evaluate the models.

We utilized the pretrained BiomedVLP-CXR-BERT-specialized architecture and weights, \cite{boecking2022making}, freezing the first 8 layers of the BERT encoder, while leaving the rest of the text encoder and vision encoder unfrozen. For each of the two models, we dropped data samples that did not contain the necessary text sections, and as a result, the Findings+Impressions model had slightly fewer training samples than the Impressions-only model.

Each model was trained for 30 epochs using the CLIP loss described in \cite{radford2021learning}. At train time, the images were augmented with a random resized crop, random affine transformation, and random color jitter before being resized to 224 by 224 and normalized. We trained for 30 epochs with a learning rate of 0.0001 and batch size of 32, validating every epoch and saving the model checkpoint that minimized CLIP loss on the validation set.

\subsection{Zero-shot classification} \label{zeroshot}
For each of the 4 clinical labels (Cardiomegaly, Edema, Consolidation, and Pleural Effusion) we provided 1-5 short captions. For all but Consolidation, these captions were taken from the query captions used in \cite{zhang2020contrastive}, while for Consolidation, the caption was hand-written based on findings within the MedPix dataset. All captions describing a particular label were passed through the text encoder, an average was computed, and then divided by its norm to produce a single query embedding corresponding to that label. This process was repeated to create a query embedding for each label.

\begin{figure}[H]
    \centering
    \includegraphics[scale=0.45]{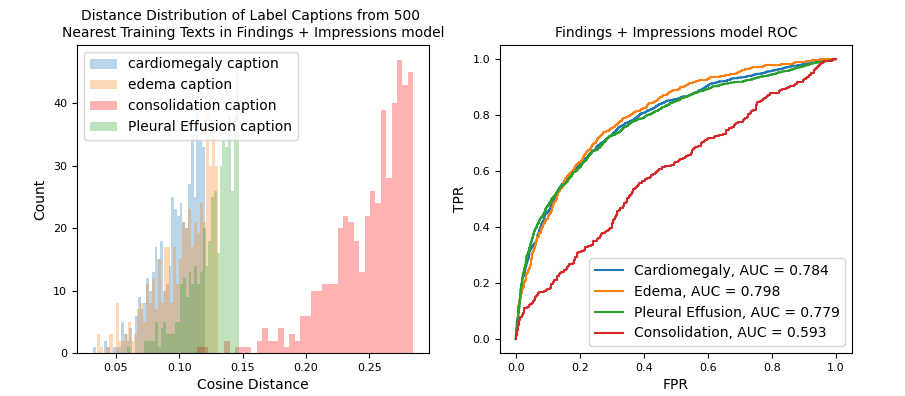}
    \caption{Left plot shows histograms of cosine distances from each label caption embedding to their 500 nearest text embeddings in the train set in the Impressions + Findings model. On the right are the ROC curves for these label captions corresponding to the zero-shot classification on the test set with pure labels. Label captions with higher KNN Distance have poorer performance on zero-shot classification as measured by the AUROC score.}
    \label{fig:ROC_KNN_IF}
\end{figure}

We constructed zero-shot classifiers as in \cite{palepu2022self} to predict the clinical labels from input images by applying the label query embeddings to our CLIP models. The input images were processed by the vision encoder to produce image embeddings, and these were divided by their norm and multiplied by the various queries to produce a cosine similarity corresponding to each label. The ROC curves shown in \ref{fig:ROC_KNN_I} and \ref{fig:ROC_KNN_IF} were produced by applying this zero-shot classifier to single-label images.

\newpage
\subsection{Query Captions for Labels} \label{query_caption}
We began with 5 captions for each of Cardiomegaly, Pleural Effusion, and Edema as provided by \cite{zhang2020contrastive}, and 2 captions that we created for consolidation, as these were not provided. We evaluated each individual caption on the training set using the Zero-shot classification procedure described in \ref{zeroshot}, and we threw out a total of 3 captions that had poor or negative AUROCs, resulting in the following 14 captions.

\begin{table}[H]
 \caption{Table Showing Query Captions Used for the Class Labels}
  \centering
  \begin{tabular}{ll}
    \toprule
    
    Class Label          & Caption \\
    \midrule
    \multirow{4}{*}{Cardiomegaly}      &Cardiomegaly is present. \\
    &The heart shadow is enlarged.\\
    &The cardiac silhouette is enlarged.\\
    &Cardiac enlargement is seen.\\\midrule
    \multirow{4}{*}{Pleural Effusion} &A pleural effusion is present.\\
    &Blunting of the costophrenic angles represents pleural effusions.\\
    &The pleural space is partially filled with fluid.\\
    &Layering pleural effusions are present.\\\midrule
    \multirow{5}{*}{Edema} &Mild interstitial pulmonary edema is present.\\
    &The presence of hazy opacity suggests interstitial pulmonary edema.\\
    &Moderate alveolar edema is present.\\
    &Mild diffuse opacity likely represents pulmonary edema.\\
    &Cardiogenic edema likely is present.\\\midrule
    \multirow{1}{*}{Consolidation} &Dense white area of right lung indicative of consolidation.\\

    \bottomrule
  \end{tabular}
  \label{tab:query_table}
\end{table}

\subsection{Additional Figures}

\begin{figure}[btp]
    \centering
    \includegraphics{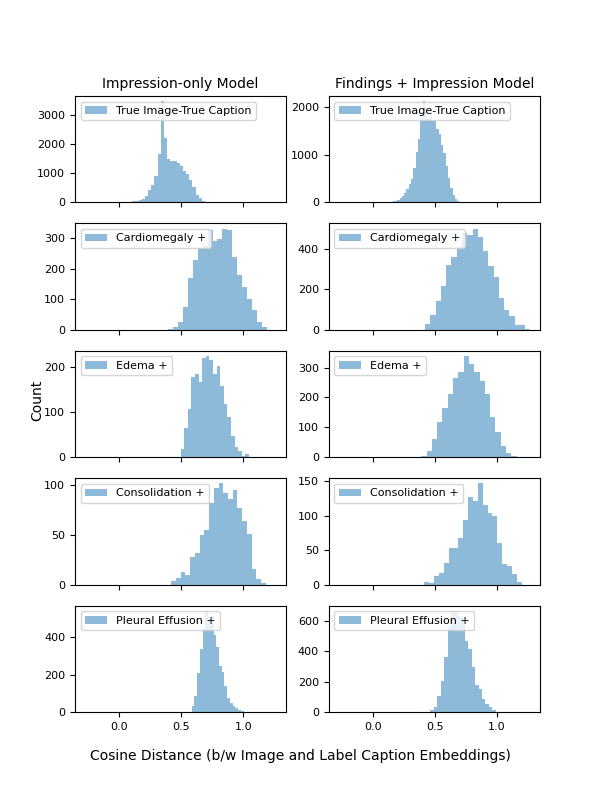}
    \caption{The top row shows the distribution of true image - true caption distances for each model. The next four rows display the cosine distances between the four label captions and test images corresponding to that label.}
    \label{fig:pos_neg_img_fig}
\end{figure}

\begin{figure}[btp]
    \centering
    \includegraphics[scale=0.6]{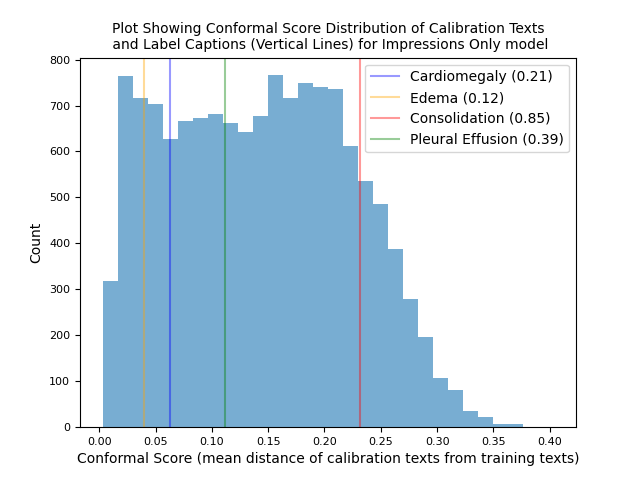}
    \caption{The plot show the conformal score distribution of calibration captions computed from the mean 500 Nearest Neighbor distance from training texts in Impressions only model. The vertical lines correspond to the score for our four label captions. The number in the bracket beside the label names in the plot indicate the coverage rate required to admit the caption corresponding to that label. A higher required coverage rate for admitting a caption in conformal prediction procedure indicates that caption for that label is less similar to the captions in training and calibration set.}
    \label{fig:knn_conformal_i}
\end{figure}

\begin{figure}[btp]
    \centering
    \includegraphics[scale=0.6]{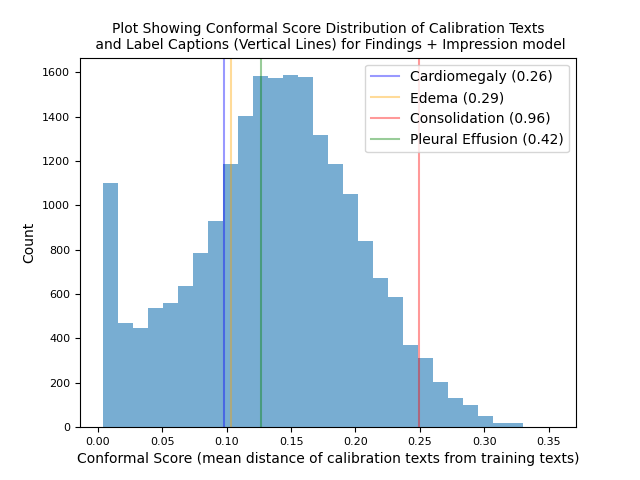}
    \caption{The plot show the conformal score distribution of calibration captions computed from the mean 500 Nearest Neighbor distance from training texts in Impressions and Findings model. The vertical lines correspond to the score for our four label captions. The number in the bracket beside the label names in the plot indicate the coverage rate required to admit the caption corresponding to that label. A higher required coverage rate for admitting a caption in conformal prediction procedure indicates that caption for that label is less similar to the captions in training and calibration set.}
    \label{fig:knn_conformal_f}
\end{figure}
\end{document}